\def\year{2019}\relax
\documentclass[letterpaper]{article} %DO NOT CHANGE THIS
\usepackage{aaai19}  %Required
\usepackage{times}  %Required
\usepackage{helvet}  %Required
\usepackage{courier}  %Required
\usepackage{url}  %Required
\usepackage{graphicx}  %Required
\usepackage{amsmath}
\usepackage{amsfonts}
\usepackage{booktabs}
\usepackage[usenames, dvipsnames]{color}
\usepackage{soul}
\usepackage{cancel}
\usepackage{amsthm}
\usepackage{hyperref}
\usepackage{diagbox} 
\usepackage{hyperref}
\usepackage{subfigure}
\usepackage{booktabs,caption}
\usepackage[flushleft]{threeparttable}
\usepackage[explicit]{titlesec}

\usepackage[lined,commentsnumbered,ruled,]{algorithm2e}

\newcommand{\vect}[1]{\boldsymbol{#1}}

\newtheorem{definition}{Definition}

\begin{document}
% The file aaai.sty is the style file for AAAI Press 
% proceedings, working notes, and technical reports.
%
\title{Structure-Preserving Transformation: Generating Diverse and Transferable Adversarial Examples %that Keep Structure Patterns
}
\author{
}
\author{Dan Peng\textsuperscript{1},
Zizhan Zheng\textsuperscript{2},
Xiaofeng Zhang\textsuperscript{1}\\
\textsuperscript{1}{Harbin Institute of Technology (Shenzhen)}\\
\textsuperscript{2}{Tulane University, USA}\\
pengdan@stu.hit.edu.cn,
zzheng3@tulane.edu,
zhangxiaofeng@hit.edu.cn}
% %\title{The Water Mill}
\maketitle

%
%\titlerunning{Abbreviated paper title}
% If the paper title is too long for the running head, you can set
% an abbreviated paper title here
%
%\author{Dan Peng\inst{1}
%Zizhan Zheng\inst{2}  \and
%Xiaofeng Zhang\inst{1} \\
%\institute{\inst{1}Harbin Institute of Technology \and \inst{2}Tulane University, USA
%}
%}
%

% First names are abbreviated in the running head.
% If there are more than two authors, 'et al.' is used.

%
\maketitle
% \blue{
% \begin{itemize}
%     \item I am not sure which property of the adversaries is more suitable for mentioning in the title, "Natural" or "Diverse"? 
%     \red{Why the proposed approach is natural and why is it diverse? Think about how to justify each of them and choose the one that is easier to justify}
%     \item What do you think we call the transformation "quasi-isometric transformation"? isometric map means it keep distance or shape after mapping. In mathematics, quasi-isometry is an equivalence relation on metric spaces that ignores their small-scale details in favor of their coarse structure. \url{https://en.wikipedia.org/wiki/Quasi-isometry}
%     \red{We can use it if we can clearly define what the metric is in our setting and how the two properties mentioned in the wikipedia article are satisfied. Otherwise, it's better to choose a term that is more intuitive such as "structure-preserving transformation"}
%     \blue{nice! then let's call it structure-preserving transformation, but what abbreviation for it, "SPT"? }\red{OK to me.}
% \end{itemize}
% }

\begin{abstract}

Adversarial examples are perturbed inputs designed to fool machine learning models. Most recent works on adversarial examples for image classification focus on directly modifying pixels with minor perturbations. A common requirement in all these works is that the malicious perturbations should be small enough (measured by an $L_p$ norm for some $p$) so that they are imperceptible to humans. However, small perturbations can be unnecessarily restrictive and limit the diversity of adversarial examples generated. Further, an $L_p$ norm based distance metric ignores important structure patterns hidden in images that are important to human perception. Consequently, even the minor perturbation introduced in recent works often makes the adversarial examples less natural to humans. More importantly, they often do not transfer well and are therefore less effective when attacking black-box models especially for those protected by a defense mechanism. In this paper,  we propose a {\it structure-preserving transformation (SPT)} for generating natural and diverse adversarial examples with extremely high transferability. The key idea of our approach is to allow perceptible deviation in adversarial examples while keeping structure patterns that are central to a human classifier. Empirical results on the MNIST and the fashion-MNIST datasets show that adversarial examples generated by our approach can easily bypass strong adversarial training. Further, they transfer well to other target models with no loss or little loss of successful attack rate. 

\end{abstract} 
\section{Introduction}

%\red{
%\begin{itemize}
%\item read carefully ``EAD: Elastic-Net Attacks to Deep Neural Networks via Adversarial Examples'' (AAAI'18) and "Towards Evaluating the Robustness of Neural Networks" (SP'17). Learn how they motivate their problems, explain their models, describe and evaluate their algorithms.
%\end{itemize}
%}

% % current problem and why investigate adversarial examples.
% \blue{Deep neural networks (DNNs) are highly expressive learning models that have achieved the start-of-the-art accuracy on a variety of complex learning tasks such as speech recognition and image classification.
% However, DNNs have been shown to be highly vulnerable to adversarial attacks that intentionally %introduce worst-case perturbations to the input. 
% perturb inputs to mislead the learning models. This raises serious concerns on the security and integrity of existing %state-of-the-art 
% DNNs in various security-sensitive applications. ~\cite{szegedy2013intriguing,2014arXiv1412.6572G}. These adversarial examples expose ``blind spots'' in DNNs and can also be used to train a more robust model that is resilient to adversarial examples, known as adversarial training \cite{2017towardsmadry}. }

Deep neural networks (DNNs) have achieved %an overwhelming 
phenomenal success in computer vision by showing superber accuracy over traditional machine learning algorithms. However, recent works have demonstrated that DNNs are vulnerable to adversarial examples that are generated for malicious purposes by slightly twisting the original images~\cite{szegedy2013intriguing,2014arXiv1412.6572G}.
%. Such small or even tiny perturbations will in turn lead the DNNs to misclassify the twisted images
This observation has raised serious concerns on the robustness of %challenge to 
the state-of-the-art DNNs and limited their applications in various security-sensitive applications. On the contrary, the adversarial examples can be aggregated to augment training datasets for learning a more robust DNN, known as adversarial training \cite{2017towardsmadry}. %\blue{-- have rewritten}

% current work mainly focus on perturbation-based attack.
% and generative-based attack encounter low transferability .

\begin{figure}[t]
\centering
\includegraphics[height=0.75 in,width=0.45\textwidth]{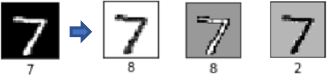}
\caption{Adversarial examples generated
by SPT. The first one is the original image selected from MNIST. The last three are adversarial examples crafted by SPT for attacking three different target classifiers, where they are classified as 8, 8, 2, respectively.}
\label{fig:intro}
\end{figure}
Generally speaking, adversarial examples can be any valid inputs to machine learning models that are intentionally designed to cause mistakes~\cite{elsayed2018adversarial}.
%\blue{In this paper, we focus on attacking machine learning models at the testing stage, where an attacker could manipulate the input data before feeding them to any machine learning algorithms so that the outputs differ from their ground truth labels.} 
% \blue{ adversarial examples are inputs to machine learning models that an attacker intentionally designed to cause the model to make a mistake} (that is, the output label differs from the ground truth). 
%\blue{adversarial examples assume that attackers could manipulate the input data before feeding them into any machine learning algorithms to maliciously change the outputs so that they differs from the human labeled ground truth labels}\blue{-- have rewritten}
For visual object recognition, we rely on human labelers to obtain the ground truth labels, which are unknown to the attacker before adversarial examples are generated. To bypass this dilemma, a common attack strategy is to start with a clean image where the ground truth label is already known and modify it so that the new image is visually similar to the original image while its output label differs from the ground truth label of the clean image. The intuition behind this approach is that by ensuring visual similarity, the two images are likely to share the same ground truth label.   
%For an image classification DNN, 
%an adversarial example is a modified version of a clean image so that the two images are visually similar but the adversarial example is misclassified by the DNN (that is, its output label differs from the ground truth label of the clean image). 

A simple approach for ensuring visual similarity that has been intensively studied in the literature is to introduce small perturbations into pixels such that the distortion between the adversarial example and the original image %is required to be small enough so that the modification 
is human imperceptible~\cite{2014arXiv1412.6572G,su2018attacking,chen2017ead,carlini2017towards}. %, which has been where the distortion is typically measured in an $L_p$ norm for some $p$~\cite{2014arXiv1412.6572G,su2018attacking,chen2017ead, carlini2016towards}. 
However, the noisy and meaningless worst-case perturbations introduced %superimposed onto original images 
by these approaches often make the generated adversarial images less natural%, that is, they are not the ones that are likely to be found in reality
~\cite{2014arXiv1412.6572G}. Moreover, perturbation-based adversarial examples exhibit low transferability, which makes them less effective for black-box attacks especially for those with a defense mechanism~\cite{kurakin2016adversarial}. There are some recent works~\cite{zhao2017generating,song2018generative} that relax the small perturbation requirement by utilizing generative networks (e.g., GANs). However, they attempt to generate adversarial examples with the identical distribution as the original dataset %but not limit to meet small perturbation requirement, 
which limits their transferability and therefore the effectiveness in black-box attacks. To improve the robustness of DNNs against adversarial examples, a large number of defense mechanisms have been proposed in the literature~\cite{kurakin2016adversarial,2017towardsmadry,guo2017countering,cisse2017parseval}. %Although inspired by different perspectives, 
However, they are mainly designed to make classifiers more robust to small perturbations. 

In this paper, we propose a {\it structure-preserving transformation (SPT)} for generating natural and diverse adversarial examples  with  extremely  high transferability.  
%In the paper, we propose a structural-preserving transformation to generate these structural-preserving while legible for human adversarial examples. 
Our approach is inspired by the fact that humans recognize object mainly from their shapes and structure patterns, supplemented by the color and brightness information. For example, consider the four images of digit 7 in Figure~\ref{fig:intro} where the first one is taken from MNIST while the next three are adversarial examples crafted by our algorithm against different target models. Despite the different background and foreground brightness in the four images, a human being can easily recognize the digit in all the cases since they share similar structure patterns. However, the last three images are mispredicted as digit 8 or 2 by DNNs. This inspires us to design adversarial examples that allow  perceptible deviation (such as color and brightness) while keeping structure patterns that are central to a human classifier.
%preserve the structure information while only changing the color and brightness information. 
%In fact, the last three images in the figure are adversarial examples crafted by our algorithm. They are mispredicted as digit 8 by DNNs. 
We show that our structure-preserving adversarial examples are highly transferable with no loss or little loss of successful rate when applied to black-box attacks even when a defense mechanism is applied. This can be explained as follows. %using substitute model, with defense or not. 
%This indicates that previous defense methods are not effective against our structural-preserving adversarial examples. 
Traditional supervised learning models including DNNs all rely on the ${\it i.i.d.}$ assumption %They can classify test data drawn from the same distribution as the training data with high accuracy. However, 
and become much less effective when test data and training data are drawn from different distributions. Current defense mechanisms do not improve that. Our approach generates adversarial examples that may have a totally different distribution from the original dataset and is therefore highly transferable.

An important limitation of previous attacks is that they can only generate a single type of adversarial examples. For perturbation-based attacks, a crafted adversarial example is a replica of the original image except for minor perturbation. That is, the attacker's action space (with respect to a single clean image) is a small neighborhood of the clean image under some metric. For generative-network-based attacks, they generate adversarial examples that obey the same distribution of the original images. That is, the attacker's action space is the set of images that follow the same distribution as the clean images. Although they have relaxed the small perturbation requirement, %more elastic similarity than that of perturbation-based attack, however, 
they still keep the same characteristic of the original images. For instance, the adversarial examples generated from MNIST images always have the same background color (black) and foreground color (white). %Therefore, current work can not generate diverse adversarial examples. 

In contrast, the attacker's action space of our approach is much larger. It includes all the images that preserve the structure of the clean image. As shown in Figure~\ref{fig:intro}, by preserving structure, our adversarial examples are natural, clean, legible to human beings. More importantly, for different target models or different initialization settings, the adversarial examples generated using our method are sufficiently diverse. They are distinct from the original images, and at the same time, they are also different from each other. For instance, the three adversarial examples shown in Figure~\ref{fig:intro} %to attack different target models 
exhibit different background and foreground gray-level (intensity), which do not follow the %and differ from the original images which obeying different 
distribution of the original dataset.
This is why our approach can generate more diverse adversarial examples, which in turn makes our approach more difficult to defend. In particular, a defender that is adversarially trained with one type of adversarial examples is still vulnerable to other types of adversarial examples.

This work broadens the scope of adversarial machine learning by showing a new class of adversarial examples that follow different distributions from the training dataset while still being legible and natural to humans. Our study reveals the weakness of current defense mechanisms in the face of structure-preserving attacks that go beyond the small perturbation constraint. 

%We hope to broaden the significance of the notion of adversarial examples, and attempt to find adversarial examples that have a different distribution from the training dataset while still being legible and reasonable for human beings. Our study %inject new insight
%sheds light on how to infer from data obeying different distributions from the training dataset and how to improve the robustness of the existing models. 

Our main contributions can be summarized as follows.
\begin{itemize}
    \item We propose a structure-preserving transformation (SPT) to generate natural and diverse adversarial examples. %with high transferability. 
    We show that an SPT can be easily trained with its learning parameters converging fast to the global optimum.
    
    \item We evaluate our approach on two datasets and demonstrate that our structure-preserving adversarial examples achieve extremely high transferability even when a defense mechanism is applied. %with no loss or little loss of successful attack rate when perform black-box attack using substitute model, with defense or not.
    
    \item We show that our structure-preserving adversarial examples can easily bypass strong adversarial training in the public MNIST challenge. The generated adversarial examples can dramatically reduce the accuracy of the white-box Madry network from 88.79\% to 9.79\% and the accuracy of the black-box Madry network from 92.76\%  to 9.80\%.   %from the best result in leader board with 88.79\% and dropping the accuracy of 
  
    %while the best result is 92.76\%.
    % \red{Ideally, you should compare your approach with other approaches you have evaluated or the state-of-the-art.}
\end{itemize}

\section{Background}

% \blue{We review some recent works on adversarial examples and defense methods. Although adversarial examples can be crafted for many domains, we focus on image classification tasks in this paper and will use the words ``examples'' and ``images'' interchangeably.}

%\blue{In this section, we briefly review most relevant works on generating adversarial examples as well as defense mechanisms proposed to thwart adversarial examples. Moreover, we analyze the difference between traditional adversarial examples and general adversarial examples with an emphasis on image classification tasks.} \blue{--have rewritten}

In this section, we briefly review recent studies on adversarial examples and defense mechanisms. We start with a simplified view of adversarial examples that has been broadly adopted in the literature and then discuss the more general definition used in this paper.

\subsection{Traditional Adversarial Examples}
In most previous works, an adversarial example is defined as a modified version of a clean image so that the two images are visually similar but the adversarial example is misclassified by the DNN (that is, its output label differs from the ground truth label of the clean image). A common requirement in many of existing studies is that the distortion between the adversarial example and the original image should be small enough so that the modification is human imperceptible. Also, the distortion is typically measured in an $L_p$ norm for some $p$. In particular, the $L_\infty$ norm measures the maximum variation of the distortions in pixel values~\cite{2014arXiv1412.6572G}, the $L_1$ and $L_2$ norms measure the total variations %sum up the variations of the introduced perturbations
~\cite{chen2017ead,carlini2017towards}, while the $L_0$ norm corresponds to the total number of modified pixels~\cite{su2018attacking}.
Moreover, psychometric perceptual similarity measures that are more consistent with human perception %than prior $L_p$ norm measurements
have also been studied~\cite{rozsa2016adversarial}.

Various techniques for crafting adversarial examples have been proposed. Most of them focus on adding small perturbations to the inputs, where several techniques have been studied. In gradient-based approaches such as the popular Fast Gradient Sign Method (FGSM)~\cite{2014arXiv1412.6572G} and many of its variants, adversarial examples are generated by superimposing small perturbations along the gradient direction (with respect to the input image). More powerful attacks can be obtained by considering multi-step variants of FGSM such as Projected Gradient Descent (PGD)~\cite{2017towardsmadry}.
%adversarial examples are generated by superimposing small gradient-based perturbation onto original images, based on the \blue{observation that small perturbations along the gradient direction (with respect to the input image) could mislead the employed DNNs~\cite{2014arXiv1412.6572G}. 
%A stronger attack called Projected Gradient Descent (PGD) \green{repeats FGSM with projected step several times, starting and re-staring with random perturbed original images, not clear}~\cite{2017towardsmadry}.\blue{--have written}
% By restart with random perturbed original images, and repeating  projection step several times. a stronger attack called Projected Gradient Descent (PGD) can be obtained~\cite{2017towardsmadry}.
 In optimization based approaches, an optimization problem is solved to identify the optimal perturbation over the original image that can lead to misclassification~\cite{carlini2017towards,szegedy2013intriguing}. More recently, generative networks including auto-encoders~\cite{baluja2017adversarial} and generative adversarial networks (GANs)~\cite{xiao2018generating} have been used to generate adversarial examples, where the small  perturbation requirement is imposed on the image space and is included in the objective function to be trained.
%  where the small perturbation requirement is either imposed on the original image space or a low-dimensional latent space and is included in the objective function to be trained. 
 %either imposed on the original image space (similar to perturbation-based methods) or a low-dimensional latent space.

\subsection{General Adversarial Examples}
Essentially, adversarial examples are inputs to machine learning models that are intentionally designed to make outputs differ from human labeled ground truth~\cite{elsayed2018adversarial}. 
There are two important points to be clarified in this definition. First, %when we talk about
adversarial examples should be %we mean these images that 
meaningful and recognizable to humans~\cite{song2018generative}. Although unrecognizable images can be generated to mislead classifiers~\cite{nguyen2015deep}, they are hard to label by humans to get the ground truth. Second, the small perturbation requirement is not a necessity. Instead, it is introduced to allow the adversarial examples share the ground truth labels of the original images to simplify the design of attacks. %Importantly, To cause a mistake in learning models, 
%for generating adversarial examples, we need to know human labeled ground truths of adversarial examples, while the ground truths are labeled manually only when the adversarial examples have been generated. 
To cause a mistake in learning models, we need to ensure that adversarial examples are classified into labels that are different from their ground truth labels, while the latter is unknown until the adversarial examples have been generated and labelled by humans. To avoid this dilemma, traditional  approaches introduced the small perturbation requirement to ensure the \emph{visual similarity} between adversarial examples and clean images so that they are likely to share the ground truth labels. %of clean images could be shared with adversarial examples. 
%In the paper, we add structure-preserving restriction between adversarial examples and clean images to ensure they can share thesame ground truth. 
In contrast, we propose to use structure-similarity to model \emph{semantic similarity} in this paper. Both perturbation-based and structure-preserving adversarial examples are consistent with the general definition of adversarial examples. 

\subsection{Defense Techniques}
To improve the robustness of machine learning models against adversarial examples, a number of defense mechanisms have been proposed in the literature. Some approaches try to merge adversarial examples into the training dataset to decrease the model susceptibility to malicious attacks~\cite{kurakin2016adversarial,2017towardsmadry}. %Alternatively,
Researcher have also considered techniques for removing adversarial perturbations during the test stage~\cite{guo2017countering} and modifying network structures to improve the model robustness~\cite{cisse2017parseval}. 
% Although inspired by different perspectives, %a shared design principle of 
However, most defense methods are ineffective to the newly proposed attacks \cite{athalye2018obfuscated,carlini2017adversarial}. Only a few state-of-the-art defense models have demonstrated their robustness to adversarial examples \cite{2017towardsmadry,samangouei2018defense}, which, however, are mainly designed for  perturbation-based attacks. 
To shed light on the weakness of existing defense mechanisms, we have evaluated our structure-preserving attacks against %on strong 
adversarial training with PGD~\cite{2017towardsmadry}, which has been demonstrated to be the most effective defense method on the MNIST dataset~\cite{athalye2018obfuscated}.

% \subsection{White-Box Attack Models}
% \textbf{One-step gradient-based approaches}\\
% \textbf{Adversarial transformation transformation}

% \subsection{Black-Box Attack Models}
% \subsection{Defense Mechanisms}
% \textbf{Feature Squeezing}\\
% \textbf{Defense-GAN}\\

%\section{Proposed Histogram Transformation}
\section{Structure-Preserving Attacks to DNNs}

% \red{
% \begin{itemize}
%     \item add a formal description of the algorithm similar to Algorithm 1 in the EAD paper.\\
%     \blue{added, please check if it makes sense }
%     \item{It seems that Algorithm 1 is a direct application of the Adam algorithm, then there is no need to include it.}
%     \blue{Just confirm, do you mean to remove Algorithm 1?}\red{Yes, I think so.}
% \end{itemize}
% }

The proposed approach is called \textbf{S}tructure-\textbf{P}reserving \textbf{T}ransformation (SPT), which attempts to transform the input images into adversarial examples against a deep neural network while keeping the structure patterns of the original images. SPT can be either untargeted or targeted and can be used for both white-box and black-box attacks. In this paper, we focus on untargeted white-box and black-box attacks.

Let $\mathbf{x} \in \mathcal{X}$ be an unlabeled image and $\mathcal{Y}$ the set of class labels. %\blue{ the element of $\mathcal{Y}$ should be  probability vector}. 
Let $f$ be the target network that outputs a probability distribution across the class labels. The target model assigns the most possible class $
C(\mathbf{x}) = \arg \max_{k \in \mathcal{Y}} [f(\mathbf{x})]_k$ to the input image $\mathbf{x}$. Let $\mathbf{x'}$ denote the crafted adversarial image of $\mathbf{x}$ using the proposed transformation. %\textbf{H}istogram \textbf{T}ransformation (HT). 
The general transformation function $g_{f,\vect{\theta}}$ can be defined as
%---------------------------------
\begin{eqnarray}
g_{f,\vect{\theta}}:  \mathbf{x} \in \mathcal{X}\to  \mathbf{x'}, 
\end{eqnarray}
% ---------------------------------
%where $g_{f,\boldsymbol{\theta}}$ is the histogram transformation function with respect to the target model $f$, 
where %$f$ is the hypothesis function of target model that outputs a probability distribution across class labels, and
$\vect{\theta}$ is the parameter vector of % the transformation 
function $g_{f,\vect{\theta}}$. Given the ground truth class label 
of the original image $\mathbf{x}$, $C_\mathrm{true}(\mathbf{x})$,
our approach tries to craft an adversarial image (transformed image $\mathbf{x'}$) to mislead classifier $f$ so that $C(\mathbf{x'}) \ne C_\mathrm{true}(\mathbf{x})$.

%\red{Something is wrong here. First, is the ``ground truth label'' the right name for $\vect{l}_\mathrm{true}(\mathbf{x})$ since it is a vector (a distribution I guess). But  if that's the case, it makes more sense to simply use $f(\mathbf{x})$. The more serious problem is that even if the two distributions are different, they may still lead to the same label}

Transformations in spatial domains are commonly used in image enhancement where an image is transformed to a new image for better visual quality by directly manipulating pixels, which can be formalized as follows.
% As one of the two broad classes of image enhancement techniques, spatial domain refers to the aggregation of the pixels composing an image. Methods in this category 
% which are based on direct manipulation of pixels in an image. 
% In this paper, 
% Formally, a spatial domain transformation can be denoted by the following expression:}
%----------------------------
\begin{eqnarray} \label{eq:spatial_proce}
\mathbf{x'}(m,n) = T_{m,n}[\mathbf{x}(m,n)]
\end{eqnarray}
%---------------------------------
where $\mathbf{x}(m,n)$ (resp. $\mathbf{x'}(m,n)$) are the intensity of the input (resp. output) image at the coordinate $(m,n)$, %$S(m,n)$ is a neighborhood of $(m,n)$, %$\vect{o}$ is the processed image, 
and $T_{m,n}$ is %an operator
a transformation on $\mathbf{x}$ that depends on $(m,n)$. Note that in general, the transformed value of any pixel may depend on all the pixels in some neighborhood of it, e.g., considering the convolution and pooling operations in CNN~\cite{krizhevsky2012imagenet}. %\blue{the "process on neighborhood" included in the transformation T ,and the variable of transformation T is point (m,n) itself} \red{This means $T$ should depend on $(m,n)$ then. I have changed it to $T_{m,n}$} %informally, the neighborhoods of point $(\mathrm{m},\mathrm{n})$ are the set of the pixel around the point $(\mathrm{m},\mathrm{n})$. 
The simplest form of $T$ is when the neighborhood of a pixel is a singleton, that is, it only contains the pixel itself. %is of size 1 x 1(that is, a single pixel). 
In this case, $\mathbf{x'}(m,n)$ depends only on $\mathbf{x}(m,n)$ and $T$ becomes a gray-level  transformation that depends only on pixel values and is independent of their coordinates.
% transformation function of the form
% %----------------------------
% \begin{eqnarray}\label{eq:single_pixel_transfor}
% s = T(r)
% \end{eqnarray}
% %---------------------------------
% where $r$ and $s$ are variables denoting, respectively, the gray level of $\vect{i}(\mathrm{m},\mathrm{n})$ and $\vect{o}(\mathrm{m},\mathrm{n})$ at any point $(\mathrm{m},\mathrm{n})$. 
% There are many transformations available for selection. Among them, 
As a special case of such singleton-based transformations, power functions %-law $s=cr^{\gamma}$ 
%are often used in gamma correction which 
are commonly used in image enhancement to allow a variety of devices to print and display images~\cite{kim1999color}. 
% In a power function based transformation, the update of intensity is totally controlled by the exponent, which typically varies between 0.04 to 25.0 as shown in Figure \ref{fig:pow_law}. 
In this paper, %Based on the above observation, 
we define the transformation function $g_{f,\vect{\theta}}$ as a linear combination of multiple power functions. For simplicity, we use $g(.)$ to denote $g_{f,\vect{\theta}}$ in the rest of this paper. Accordingly, the general form of $g(.)$ is given as, 
%---------------------------------
\begin{eqnarray}\label{eq:SPT}
% \begin{split}
g(\mathbf{x}) = \mathrm{sigmoid}(\sum_{i} w_i\mathbf{x}^{\gamma_{i}})
%\blue{removed "regularization"}
\end{eqnarray}
%----------------------------------
where {$\gamma_{i}$} is the exponent of the $i$-th basis power function. In general, these exponents can be either learned from data or chosen based on domain knowledge and fixed in advance. 
For simplicity, we manually choose the values of {$\gamma_{i}$}'s in this paper. %rather than learn them during training. 
The weight $w_{i}$ is a scalar shared with all the elements in $\mathbf{x}^{\gamma_{i}}$. These weights $\{w_{i}\}$ are the parameters to be learned. That is, $\vect{\theta}=\{w_{i}\}$. 
% Moreover, an $L_2$ regularization term is introduced in~\eqref{eq:SPT}, which is used to penalize intensely dramatic image transformations that may deteriorate the visual quality of images. $\alpha$ is a scalar parameter that controls the strength of regularization.
The $\mathrm{sigmoid}$ function is adopted to further guarantee the range of output falls into [0,1].

%Formally, we give a definition of structure pattern mentioned above.
A notable property of any singleton-based transformation including ours is that all the pixels with the same gray level in the original image have the same gray level in the transformed image. Further, our transformation is typically injective, i.e., it ensures that pixels with different gray levels in the original image typically have different gray levels in the transformed image.  
That is, such a transformation keeps the {\it structure patterns} of images, which is formally defined as follows.
\begin{definition} \label{def:structure_pattern}
\textbf{Structure Pattern} The set of all pixels with the same grey-level in an image is called a structure pattern of the image. Symbolically, given an image $\mathbf{x}$, the set $\{(m,n) \lvert \mathbf{x}(m,n) = c\}$ is called a structure pattern with gray level $c$. 
\end{definition}

\subsection{Training}
SPT is trained together with the target model by adding an extra layer in front of the target network. For untargetd attacks, our objective is to find the parameters $\vect{\theta}$ so that $C(\mathbf{x'}) \ne C_\mathrm{true}(\mathbf{x})$. 
To improve the chance of successful attacks, we aim to find $\vect{\theta}$ so that the distance between the predicted logits (that is, the output of the classifier $f(\mathbf{x})$) and the one-hot encoding of ground truth class on all the training data is maximized. 
% For an input image $\mathbf{x}$,its ground truth logit $l_\mathrm{true}(\mathbf{x})$ is defined as the one-hot encoding of the ground truth class $C_\mathrm{true}(\mathbf{x})$. That is, $l_\mathrm{true}(\mathbf{x})_k = 1$ for $k = C_\mathrm{true}(\mathbf{x})$ and $l_\mathrm{true}(\mathbf{x})_k = 0$ otherwise.
We consider the following objective function: 
%---------------------------------
\begin{eqnarray}\label{eq:obj}
\underset{\vect{\theta}}{\mathrm{argmax}}\sum_{\mathbf{x} \in \mathcal{X}} L(f(g(\mathbf{x})),l_\mathrm{true}(\mathbf{x}))) - \alpha\sum_{i}w_i^2 ,
\end{eqnarray}
%---------------------------------
where $L$ is a loss function that measures the difference between the output logit $f(g(\mathbf{x}))$ when the target model $f$ is applied to the crafted adversarial example $g(\mathbf{x})$ and the one-hot encoding of ground truth class of the original image. %\blue{added regularization} 
Moreover, an % \blue{negative?}
$L_2$ regularization term is introduced in~\eqref{eq:obj}, which is used to penalize intensely dramatic image transformations that may deteriorate the visual quality of images. $\alpha$ is a scalar parameter that controls the strength of regularization. 
Unlike most other approaches, the generated $\mathbf{x}'$ is not restricted to be similar enough to the original $\mathbf{x}$. Instead, our approach guarantees the perceptual similarity in human vision in terms of structure patterns. Hence, the corresponding objective function could be defined in a concise manner and is thus easy to converge. 

In this paper, we define the loss function $L$ as the cross entropy between the one-hot encoding ground truth class $l_\mathrm{true}(\mathbf{x})$ of the original image and the predicted logit $f(g(\mathbf{x}))$ of the corresponding adversarial image. %Without ambiguity, we use $\vect{l}(\mathbf{x})$ to denote $\vect{l}_\mathrm{true}(\mathbf{x})$.
% Formally, 
% %---------------------------------
% \begin{eqnarray}\label{eq:loss}
% L(f(g(\mathbf{x})),l_\mathrm{true}(\mathbf{x})) =- {l_\mathrm{true}(\mathbf{x})}^{T}\log(f(g(\mathbf{x}))). 
% \end{eqnarray}
% %---------------------------------
% where $T$ denotes the transpose of a vector. 
For untargted attacks, 
we then solve the following minimization problem to find the parameters $\theta$ of $g(\cdot)$:
%---------------------------------
\begin{eqnarray}\label{eq:paraEst}
\underset{\vect{\theta}}{\mathrm{argmin}}\sum_{\mathbf{x} \in \mathcal{X}}{l_\mathrm{true}(\mathbf{x})}^{T}\log(f(g(\mathbf{x}))) + \alpha\sum_{i}w_i^2.
\end{eqnarray}
%---------------------------------
%Note, we omit the negative symbol, so we turn $argmax$ operation into $argmin$ operation. 
%\blue{removed k, transported $\vect{l}$ }
Although we focus on untargeted attacks in this work, our approach extends to targeted attacks by considering a slightly different objective function:
%---------------------------------
\begin{eqnarray}\label{eq:obj_target}
\underset{\vect{\theta}}{\mathrm{argmin}}\sum_{\mathbf{x} \in \mathcal{X}} L(f(g(\mathbf{x})),l_\mathrm{target}(\mathbf{x'}))) + \alpha\sum_{i}w_i^2,
\end{eqnarray} 
%---------------------------------
where $l_\mathrm{target}(\mathbf{x'})$ is the vectorized representation of target class we want to mispredict, which can be simply defined as the one-hot encoding of the target class similar to the untargeted case. In addition to this simple encoding scheme and the cross-entropy based loss function, other encoding techniques and loss functions~\cite{baluja2017adversarial,carlini2017towards} can also be applied to our framework for both untargeted attacks and target attacks, which will be studied in the future.

To find the learning parameters $\{w_{i}\}$, we solve the above optimization problem using the Adam method~\cite{kingma2014adam} with a learning rate of $10^{-4}$. Due to the good convergence property of the cross entropy loss function, one epoch training is sufficient for all the training data. Moreover, because a small number of parameters $\{w_{i}\}$ (about 10 parameters) need to be optimized, it only takes a few minutes to train.

\subsection{Inference} 
After training a Structure-Preserving Transformation, we generate adversarial examples by performing the trained SPT on the original images. The process is fast and only involves computing a linear combination of power functions and does not require any information of the target models.

\section{Experiment Results}
% \red{we may need to remove some results/discussions to save space}
% 
% \green{you'd better organize this section following the standard one used in AAAI. Feel free to re-organize them if necessary.\\
% 1. illustrate dataset.\\
% 2. Evaluation Metrics.\\
% 3. Baseline Methods.}

% \begin{itemize}
%     \item \red{Is it useful to consider more datasets, e.g., CIFAR10 and ImagetNet?}\\
%     \blue{I have done experiments on the CIFAR10.   The biggest problem is image distortion after histogram transformation on color image, even i did transformation in YUV color space rather than original RGB space. I have no idea how to avoid this problem at present.}
%     \item \red{it helps to consider some defense algorithms }\\
%     \blue{I will complement experiments with two state-of-the-art defense methods as quickly as possible: defense-GAN \cite{samangouei2018defense} and a robust adversarial training method \cite{2017towardsmadry}}
    
% \end{itemize}

\begin{table*}[t]
\caption{\textbf{White-box attacks}. Classification accuracy of different models on the MNIST (top) and F-MNIST (bottom) datasets. In each row, the best result is highlighted in bold and the second-best result is underlined. 
%under various attack strategies under white-box scenario on%, under no defense, Defense-GAN, and PGD adversarial training.
}

\label{tab:white_attack}
\centering
\begin{tabular}{l || c |c | c| c | c|c}
% \toprule
%----------------------------------mnist----------------------------
\multicolumn{1}{c}{Defense}&
\multicolumn{1}{c}{\begin{tabular}{@{}c@{}}Target \end{tabular}} &
\multicolumn{1}{c}{\begin{tabular}{@{}c@{}}No  Attack\end{tabular}} &
\multicolumn{1}{c}{\begin{tabular}{@{}c@{}}FGSM\end{tabular}}&
\multicolumn{1}{c}{\begin{tabular}{@{}c@{}}  PGD \end{tabular}}& 
\multicolumn{1}{c}{\begin{tabular}{@{}c@{}}  C\&W\end{tabular}} & \multicolumn{1}{c}{\begin{tabular}{@{}c@{}}\textbf{SPT}\end{tabular}} \\

\hline
\begin{tabular}{@{}c@{}}No \\ Defense\end{tabular}&
\begin{tabular}{@{}c@{}} $C_{p}$\\ $C_{a0}$ \\ $C_{a1}$\\ $C_{a2}$\\$C_{a3}$ \end{tabular}&
\begin{tabular}{@{}c@{}}99.02\% \\ 98.83\% \\ 98.73\%\\ 98.33\%\\98.58\% \end{tabular}&
\begin{tabular}{@{}c@{}} \underline{9.28\%}\\ \underline{5.55\%} \\ 7.18\%\\ 8.25\%\\11.44\% \end{tabular}&
\begin{tabular}{@{}c@{}} \textbf{0.00\%}\\ \textbf{0.00\%} \\ \underline{0.03\%}\\ \underline{0.09\%} \\ \textbf{0.00\%} \end{tabular}&
\begin{tabular}{@{}c@{}} \textbf{0.00\%} \\ \textbf{0.00\%} \\ \textbf{0.00\%}\\ \textbf{0.00\%}\\\textbf{0.00\%} \end{tabular}&
\begin{tabular}{@{}c@{}} 9.74\%\\ 9.75\% \\ 10.52\%\\ 8.93\% \\ \underline{9.91\%} \end{tabular}
\\

% \hline
% \begin{tabular}{@{}c@{}}Defense- \\GAN-Rec \\ $L = 200$ \\ $R = 10 $\end{tabular}&\begin{tabular}{@{}c@{}} p\\ a0 \\ a1\\ a2\\a3 \end{tabular}&\begin{tabular}{@{}c@{}}99.02\% \\ 98.83\% \\ 98.73\%\\ 98.33\%\\98.58\% \end{tabular}&&&&\\

\hline
\begin{tabular}{@{}c@{}}  PGD Adv. Tr.\\(Tr. Acc) \\ $\epsilon$ = 0.3 \\ $\alpha$ = 0.01 \end{tabular}&
\begin{tabular}{@{}c@{}} $C_{p}$\\ $C_{a0}$ \\ $C_{a1}$\\ $C_{a2}$\\$C_{a3}$ \end{tabular}&
\begin{tabular}{@{}c@{}} 98.08\% \\ 97.65\% \\ 98.13\%\\ 98.20\%\\96.90\% \end{tabular}&
\begin{tabular}{@{}c@{}} 93.24\% \\ 84.51\% \\ 89.01\%\\ 87.78\%\\92.54\% \end{tabular}&
\begin{tabular}{@{}c@{}} 88.14\% \\ 68.07\% \\ 73.92\%\\ 73.16\%\\87.15\% \end{tabular}&
\begin{tabular}{@{}c@{}} \underline{32.00\%} \\ \textbf{5.00\%} \\ \underline{6.00\%}\\ \textbf{4.00\%}\\ \underline{45.00\%} \end{tabular}&
\begin{tabular}{@{}c@{}} \textbf{13.47\%} \\ \underline{5.10\%}\\ \textbf{4.39\%}\\ \underline{8.98\%} \\ \textbf{7.32\%} \end{tabular}\\

%---------------------------------------mnist---------------------------------------
\multicolumn{6}{c}{}\\
%------------------------------------f-mnist----------------------------------------
% \multicolumn{1}{c}{Defense}&
% \multicolumn{1}{c}{\begin{tabular}{@{}c@{}}Target \\ Classifier\end{tabular}} &
% \multicolumn{1}{c}{\begin{tabular}{@{}c@{}}No \\ Attack\end{tabular}} &
% \multicolumn{1}{c}{\begin{tabular}{@{}c@{}}FGSM\end{tabular}}&
% \multicolumn{1}{c}{\begin{tabular}{@{}c@{}}  PGD\\ $\epsilon$ = 0.3 \\ $\alpha$ = 0.01 \end{tabular}}& \multicolumn{1}{c}{\begin{tabular}{@{}c@{}}  C\&W\\ $\ell_{2}$ norm\end{tabular}} & \multicolumn{1}{c}{\begin{tabular}{@{}c@{}}\textbf{SPT}\end{tabular}} \\

\hline
\begin{tabular}{@{}c@{}}No \\ Defense\end{tabular}&
\begin{tabular}{@{}c@{}} $C_{p}$\\ $C_{a0}$ \\ $C_{a1}$\\ $C_{a2}$ \\ $C_{a3}$ \end{tabular}& 
\begin{tabular}{@{}c@{}}91.18\% \\ 91.35\% \\ 90.67\%\\ 89.79\%\\91.24\% \end{tabular}&
\begin{tabular}{@{}c@{}}\underline{7.73\%} \\ \underline{7.42\%} \\ \underline{7.88\%}\\ 9.61\%\\ \underline{4.42\%} \end{tabular}&
\begin{tabular}{@{}c@{}}\textbf{0.00\%} \\ \textbf{0.00\%} \\ \textbf{0.00\%}\\ \underline{0.08\%}\\ \textbf{0.00\%} \end{tabular}&
\begin{tabular}{@{}c@{}}\textbf{0.00\%} \\ \textbf{0.00\%} \\ \textbf{0.00\%}\\ \textbf{0.00\%}\\ \textbf{0.00\%} \end{tabular}&
\begin{tabular}{@{}c@{}}10.00\% \\ 9.18\% \\ 10.45\%\\ 10.04\%\\10.06\%\end{tabular}\\

% \hline
% \begin{tabular}{@{}c@{}}Defense- \\GAN-Rec \\ $L = 200$ \\ $R = 10 $\end{tabular}&
% \begin{tabular}{@{}c@{}} p\\ a0 \\ a1\\ a2\\a3 \end{tabular}&
% \begin{tabular}{@{}c@{}}91.18\% \\ 91.35\% \\ 90.67\%\\ 89.79\%\\91.24\% \end{tabular}&&&&\\

\hline 
\begin{tabular}{@{}c@{}}  PGD Adv. Tr.\\(Tr. Acc) \\ $\epsilon$ = 0.3  \\ $\alpha$ = 0.01 \end{tabular}&
\begin{tabular}{@{}c@{}} $C_{p}$\\ $C_{a0}$ \\ $C_{a1}$\\ $C_{a2}$\\$C_{a3}$ \end{tabular}&
\begin{tabular}{@{}c@{}}74.80\% \\ 72.40\% \\ 73.27\%\\ 77.29\%\\71.47\% \end{tabular}&
\begin{tabular}{@{}c@{}}68.92\% \\ 63.17\%\\ 65.18\% \\ 74.52\%\\ 64.04\%\end{tabular}&
\begin{tabular}{@{}c@{}}\underline{55.40\%} \\ \underline{50.54\%} \\ \underline{51.75\%}\\ \underline{66.41\%}\\ \underline{49.34\%} \end{tabular}&
\begin{tabular}{@{}c@{}}69.00\% \\ 78.00\% \\ 77.00\%\\ 78.00\%\\66.00\% \end{tabular}&
\begin{tabular}{@{}c@{}}\textbf{10.00\%} \\ \textbf{7.24\%} \\ \textbf{9.99\%}\\ \textbf{5.71\%}\\ \textbf{9.45\%} \end{tabular}\\
\hline

\end{tabular}
\vspace{-4mm}
\end{table*}
To evaluate the performance of SPT, we compare it with three baseline attack algorithms, FGSM~\cite{2014arXiv1412.6572G}, PGD~\cite{2017towardsmadry}, and C\&W~\cite{carlini2017towards} on two popular image classification datasets, MNIST~\cite{lecun1998gradient} and Fashion-MNIST (F-MNIST)~\cite{xiao2017fashion}. %are evaluated on these datasets. %for performance evaluation. 
Both datasets consist of 60,000 training images and 10,000 testing images from 10 classes.
As in most related works, classification accuracy is used as the evaluation criterion. A stronger attack method has a lower classification accuracy. %should be as small as possible. 
% Note that the successful rate adopted in \cite{chen2017ead} is essentially complementary to accuracy metric.
Moreover, the attack methods are evaluated both on the original models as well as when a defense mechanism is applied. In this paper, adversarial training with PGD is chosen as the defense method, which has been shown to be the most effective defense method on the MNIST dataset~\cite{athalye2018obfuscated}.
% two widely adopted defense mechanisms are considered in the experiments which as \textit{no defense} and \textit{strong adversarial training} mechanism. The latter one refers to train target model with PGD. 
% \green{not clear, why pgd is strong, why should adopt}. %Two different attack modes are then separately evaluated to demonstrate the efficacy of the proposed approach. Through this empirical study, we expect the proposed SPT can drop the accuracy of target model
% achieve a better successful attack rate for two commonly seen attack modes, and meanwhile can generate a more diverse but natural distorted adversarial images. 
Evaluations in both white-box and black-box attack settings show that the proposed SPT can generate more diverse and naturally distorted adversarial images with high transferability. 

\subsection{Experiment Settings}\label{sec: setup}

\smallskip
\noindent{\bf Baseline Attacks:} As aforementioned, three baseline attack approaches are implemented for performance comparison. \begin{itemize}
    %  \item \textbf{FGSM:} The Fast Gradient Sign Method (FGSM) attack is originally proposed in \cite{2014arXiv1412.6572G}, where small gradient-based perturbations are superimposed onto the original images to mislead the classification models.
     \item \textbf{FGSM attack:} 
     %The Fast Gradient Sign Method (FGSM) attack
     FGSM superimposes small gradient-based perturbation to the original images to mislead the classification models~\cite{2014arXiv1412.6572G}.
    
    % \item  \textbf{C\&W attack:} This attack generates adversarial examples by adding $L_2$ norm based perturbations to the original images~\cite{carlini2017towards}. An optimization problem is solved to find the adversarial examples that drop the accuracy of target model as much as possible.
    
    \item \textbf{PGD attack:} 
    %The Projected Gradient Descent (PGD) attack
    It is a variant of iterative FGSM. %\blue{%It performs FGSM multiple times where 
    In each step, PGD restarts with a randomly perturbed version of the previous result~\cite{2017towardsmadry}.
    % and identifies its gradient. A small gradient-based perturbation is then added to the previous result followed by a projection.
    %}and randomly restarts throughout the attack \red{hard to understand}}.  
    % randomly start perturbing the original images several times 
    % \textit{to generate more diverse adversarial examples}. 
    
    \item  \textbf{C\&W attack:} This attack generates adversarial examples by solving $L_2 $ norm based optimization problem to find the adversarial examples that drop the accuracy of target model as much as possible~\cite{carlini2017towards}.

\end{itemize}

% \smallskip
% \noindent{\bf Baseline Attacks:} As aforementioned, three baseline attack approaches are implemented for performance comparison.
% The Fast Gradient Sign Method (\textbf{FGSM}) superimposes small gradient-based perturbation to the  original images to mislead the classification models~\cite{2014arXiv1412.6572G}; The Projected Gradient Descent (\textbf{PGD}) attack~\cite{2017towardsmadry} is a variant of iterative FGSM; \textbf{C\&W}~\cite{carlini2017towards} generates adversarial examples by solving $L_2 $ norm based optimization problem to find the adversarial examples that drop the accuracy of target model as much as possible.

\smallskip
\noindent{\bf Parameter settings:} %The experimental settings are as follows.
%\blue{%Empirically, 
For SPT, we consider a linear combination of 11 power functions with exponents $\gamma$ equal to $0.04, 0.10,0.20, 0.40, 0.67,1.0, 1.5, 2.5, 5.0,10.0, 25.0$, respectively. %their different ability to distort original image, as shown in Figure \ref{fig:pow_law}. 
The larger the $\gamma$ is, the darker the transformed image will be. The intuition behind choosing these exponents is further explained in the 
%Appendix.
our technical report~\cite{SPT}. 
%otherwise, the image will more lighter.
%} 
%the MNIST dataset has few grey levels
Since most pixel values in the MNIST dataset are close to black or white, even a large $\gamma$ will not significantly distort an image, the coefficient $\alpha$ of the penalty term in \eqref{eq:SPT} is to set to 0 for experiments on the MNIST dataset.   %even if violent transformation~\red{what is a violent transformaton?} is introduced, and thus $\alpha$ will not be applied. 
We set $\alpha = 0.6$ for experiments on fashion-MNIST dataset to avoid low image contrast for large $\gamma$ values.%due to violent transformation. 
% For the C\&W attack, we follow the settings illustrated in \cite{chen2017ead} and both the SPT and the C\&W are trained using ADAM optimizer \cite{} with the learning rate as $10^ {-4} $ to find the successful adversarial examples. 

%Different from the rest approaches, the SPT only takes one iteration step to converge which is discussed in previous section.
Both C\&W and FGSM are implemented using the public CleverHans package with the same configurations~\cite{papernot2016cleverhans}. PGD is implemented using the source code from the MNIST challenge~\cite{mnist2017challenge}.%In the implementation, 
% where the step size is set as $\alpha = 0.01$ and the value of each pixel is modified by at most 0.3 (i.e., $\epsilon = 0.3$). %control parameter $ $\alpha = 0.01$ to generate the seed image for the later perturbation.  
% 
%Our implementation is based on TensorFlow\cite{abadi2016tensorflow} and builds on open-source software: CleverHans \cite{papernot2016cleverhans} and public MNIST challenge \href{https://github.com/MadryLab/mnist_challenge}. We use machines equipped with two pieces of NVIDIA GeForce GTX 1080 GPUs.

\smallskip
\noindent{\bf Attack modes:} %Generally, the attack mode can be classified into two-fold, i.e., \textit{\textbf{White-Box Attack}} and \textit{\textbf{Black-Box Attack}}. 
Two attack modes are considered in the evaluations.
i.e., \textbf{\textit{white-box attack}} and \textbf{\textit{black-box attack}}. For \textbf{\textit{white-box attack}}, details of the target model are publicly known to the malicious parties including model structure and learned weights, datasets, and the adopted defense mechanisms. For \textbf{\textit{black-box attack}}, only the training and testing datasets could be manipulated but the rest keeps unknown to the adversaries.

\subsection{Evaluation Results for White-Box Attacks}

%\vspace{-1.4cm}
\begin{figure*}[t]  
    \includegraphics[height=2.25 in,width=0.5\textwidth]{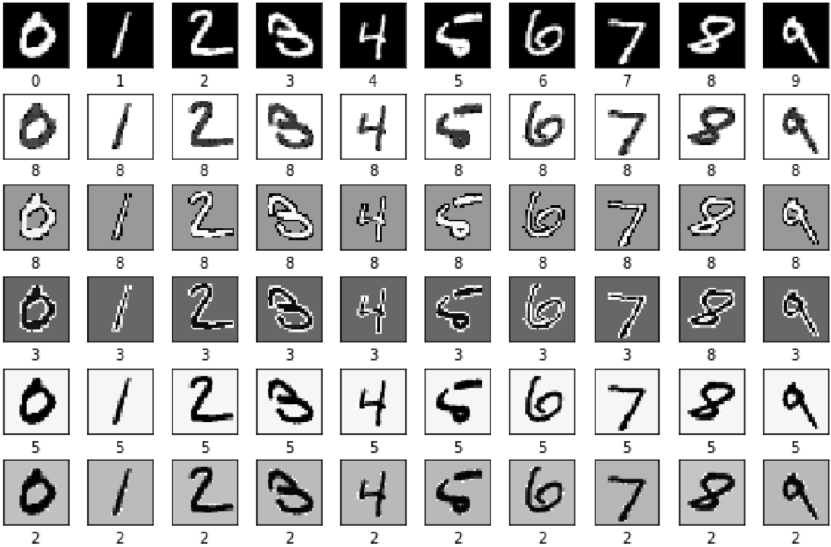} 
    \hspace{10px}
    \includegraphics{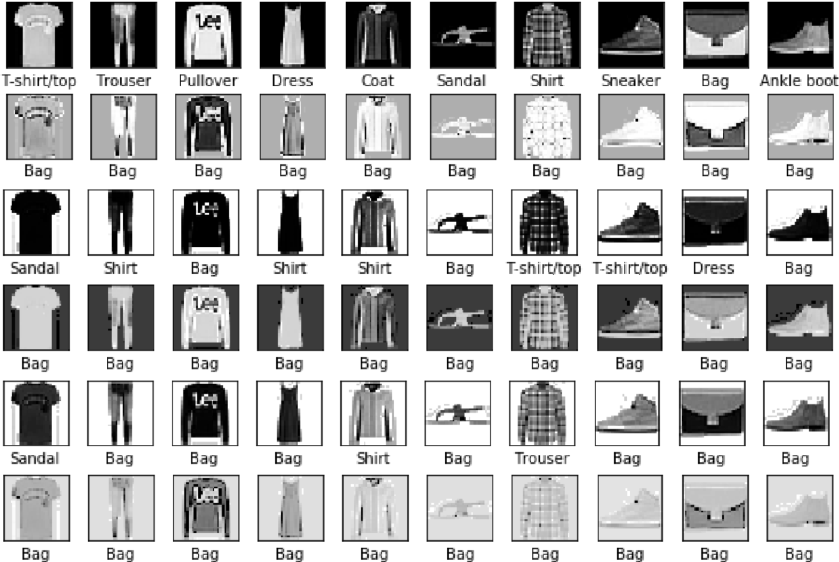}
    \caption{Visual illustration of adversarial examples generated by SPT. The first row shows the original images selected from MNIST (left) and F-MNIST (right), respectively. The other five rows show the adversarial examples crafted by SPT for each of the five classifiers, respectively.}
    \label{fig:example}
\end{figure*}

In this section, we present the experimental results in the white-box setting. 
Following the same network architecture in~\cite{baluja2017adversarial}, 
we train five networks as target models on the MNIST training dataset and F-MNIST training dataset, respectively. Each network is a combination of convolutional and fully connected layers.
%and details are provided in the Appendix.
Details are provided in our online technical report~\cite{SPT}. 
% \red{Are you saying that classifiers $p$, $a_0$ to $a_4$ are all taken from~\cite{baluja2017adversarial}? Even in that case, I think it is important to briefly described them to make the paper self-contained. Also, in the tables, it's better to call them $C_p, C_{a0}$ etc. as in~\cite{baluja2017adversarial}.}
% Each network has a mix of convolution (Conv) and Fully Connected (FC) layers, and the architecture as shown in \ref{tab:acrchiteture}. 
In Table \ref{tab:white_attack}, we report the classification accuracy of these models under various attack-defense configurations. On the attacker side, we consider five cases: when there is no attack and when one of the four attack methods is applied. % after multiple white-box attack methods against the five target models 
On the defender side, two cases are considered: when there is no defense and when PGD-based adversarial training is applied. We observe that SPT achieves low accuracy (about $10\% $ or lower) across all the scenarios. %steadily whether applying defense or not. 
When no defense is applied, both C\&W and PGD achieve very low accuracy (close to 0), while SPT is still comparable to FGSM. When PGD-based adversarial training is applied, SPT obtains overwhelming low accuracy than all the baseline attacks. These results demonstrate the effectiveness of SPT even in the presence of strong defense. %can achieve effective white-box attacks against robust target model with defense or not.
% The machine learning  model including DNNs rely on the i.i.d assumption, which achieves high correct rate for image classifation. However, they exhibits limited ability on classifying data which draw from different distribution with training data. Current defense mechanisms still suffer from the dilemma. Our SPT adversarial examples obey different distribution with original (training) images, which confer SPT adversarial examples extremely high and stable attack success rate against target models, even whit defense.

We further evaluate SPT adversarial examples on the most competitive public MNIST challenge where researchers try their best to attack the well-trained robust Madry networks~\cite{mnist2017challenge}.
% \blue{, where have trained an available robust network with PGD adversarial training \cite{2017towardsmadry} for white- and black-box attack. The objective of participants is to find a set of adversarial examples on which this network achieves only low accuracy.} 
% \red{briefly describe the challenge. why the result is different from Table 1 for SPT?} \blue{In mnist challenge, the madry network have been trained which is different from all of trained $C_{i}$}
% which has trained available robust network with PGD adversarial training.
% For comparison, we simply republish the best public result. 
%Since SPT does not enforce the small perturbation requirement, it is a bit unfair to directly compare SPT with other public results on the leaderboard. However, it gives us a intuitive feeling about the attack strength of SPT. 
% The results are summarized in Table~\ref{tab:mnist_challenge}.
Remarkably, SPT drops the accuracy of the white-box Madry network to $9.79\%$ , which is far below the current best result of 88.79\% on the leaderboard~\cite{zheng2018distributionally}. We acknowledge that the comparison is a bit unfair since SPT does not have the small perturbation requirement while the public results in the leaderboard do. However, the observed big gap does indicate the strength of SPT.
% The current work \cite{song2018generative} report a low accuracy ($15\%$) against Madry network without small perturbation restriction. However, they generate adversarial examples obeying identical distribution of original dataset, which cause the low transferability across other models.

\subsection{Black-Box Attacks and Transferability}

% \begin{table*}[t]
% \centering
% \caption{black attack on MNIST}
% \label{tab:prop_pred}
% \begin{center}
% \begin{tabular}{c c c c c c c c}
% \toprule
% Approach & $classifier_{p}^{*}$ & $classifier_{a0}$  & $classifier_{a1}$ & $classifier_{a2}$ & $classifier_{a3}$ &$classifier{a0}$+defense-GAN & $classifier{a0}$+Feature squeezing\\
% \hline
% HT & ${9.74\%}_{*}$ & 9.74\%(\%) &  10.10\%(\%) &  8.92\%(\%)&  10.17\%(\%)  &&\\

% PGD &  0.00\%&  29.24\%&  45.81\%&  35.74\%&  29.51\% & & \\
% FGSM & ${9.42\%}^{*}$& 27.95\%&  41.57\%&  30.35\%& 28.78\% && \\
% CW &  ${0.00\%}^{*}$\%& 89.00\% & 88.00\%& 89.00\%& 87.00\%&  &  \\
% \bottomrule
% \end{tabular}
% \end{center}
% \end{table*}

% \begin{table*}[t]
% \centering
% \caption{black attack on fashion-MNIST}
% \label{tab:prop_pred}
% \begin{center}
% \begin{tabular}{l c c c c c c c}
% \toprule
% Approach & $classifier_{p}^{*}$ & $classifier_{a0}$  & $classifier_{a1}$ & $classifier_{a2}$ & $classifier_{a3}$ & defense-GAN & Feature Squeezing\\
% \hline
% HT & ${10.00\%}^{*}$ & 10.02\%(99.91\%) &  10.00\%(100\%) &  13.36\%(81.65\%)&  10.00\%(100\%) && \\

% PGD &  \%&  \%&  \%&  \%&  \%  &&\\

% FGSM &  ${7.73\%}^{*}$& 11.19\%& 11.23 \%& 15.67 \%& 11.72 \% && \\
% CW &  ${0.00\%}_{*}$\%& 88.00\% & 90.00\%& 89.00\%& 90.00\%&  \% &  \\
% \bottomrule
% \end{tabular}
% \end{center}
% \end{table*}

\begin{table*}[!t]
\caption{\textbf{Black-box attacks}. Classification accuracy %using various attack strategies under white-box scenario 
on the MNIST (top) and F-MNIST (bottom) datasets where $C_{p}$ is used as the substitute model and the other models are target models. In each row, the best result is highlighted in bold and the second-best result is underlined.  %combinations. %under no defense, Defense-GAN, and PGD adversarial training.
}

\label{tab:black_attack}
\centering
\begin{tabular}{l || c |c | c| c | c|c }
% \toprule
%----------------------------------mnist----------------------------
\multicolumn{1}{c}{Defense}&
\multicolumn{1}{c}{\begin{tabular}{@{}c@{}}Target \end{tabular}} &
\multicolumn{1}{c}{\begin{tabular}{@{}c@{}}No Attack\end{tabular}} &
\multicolumn{1}{c}{\begin{tabular}{@{}c@{}}FGSM\end{tabular}}&
\multicolumn{1}{c}{\begin{tabular}{@{}c@{}}  PGD \end{tabular}}& 
\multicolumn{1}{c}{\begin{tabular}{@{}c@{}}  C\&W\end{tabular}} & \multicolumn{1}{c}{\begin{tabular}{@{}c@{}}\textbf{SPT}\end{tabular}} \\
% \multicolumn{1}{c}{\begin{tabular}{@{}c@{}} class-preserving \\ proportion\end{tabular}} \\

\hline
\begin{tabular}{@{}c@{}}No \\ Defense\end{tabular}&\begin{tabular}{@{}c@{}} $C_{p}$\\ $C_{a0}$ \\ $C_{a1}$\\ $C_{a2}$\\$C_{a3}$ \end{tabular}&
\begin{tabular}{@{}c@{}}$99.02\%^{*} $\\ 98.83\% \\ 98.73\%\\ 98.33\%\\98.58\% \end{tabular}&
\begin{tabular}{@{}c@{}} $9.42\%^{*}$\\ \underline{27.95\%} \\ \underline{41.57\%}\\ \underline{30.35\%}\\\underline{28.78\%} \end{tabular}&
\begin{tabular}{@{}c@{}} $0.00\%^{*}$\\ 29.24\% \\ 45.81\%\\ 35.74\%\\29.51\% \end{tabular}&
\begin{tabular}{@{}c@{}} $0.00\%^{*}$ \\ 89.00\% \\ 88.00\%\\ 89.00\%\\87.00\% \end{tabular}&
\begin{tabular}{@{}c@{}} $9.74\%^{*}$\\ \textbf{9.74\%} \\ \textbf{11.72\%}\\ \textbf{8.92\%} \\ \textbf{10.17\%} \end{tabular}
\\

% \hline
% \begin{tabular}{@{}c@{}}Defense- \\GAN-Rec \\ $L = 200$ \\ $R = 10 $\end{tabular}&\begin{tabular}{@{}c@{}} p\\ a0 \\ a1\\ a2\\a3 \end{tabular}&\begin{tabular}{@{}c@{}}99.02\% \\ 98.83\% \\ 98.73\%\\ 98.33\%\\98.58\% \end{tabular}&&&&\\

\hline
\begin{tabular}{@{}c@{}}  PGD Adv. Tr.\\(Tr. Acc) \\ $\epsilon$ = 0.3 \\ $\alpha$ = 0.01 \end{tabular}&
\begin{tabular}{@{}c@{}} $C_{p}$\\ $C_{a0}$ \\ $C_{a1}$\\ $C_{a2}$\\$C_{a3}$ \end{tabular}&
\begin{tabular}{@{}c@{}} 98.08\% \\ 97.65\% \\ 98.13\%\\ 98.20\%\\96.90\% \end{tabular}&
\begin{tabular}{@{}c@{}} 94.50\% \\ \underline{87.38\%} \\ 90.19\%\\ \underline{85.67\%}\\92.86\% \end{tabular}&
\begin{tabular}{@{}c@{}} 95.41\% \\ 90.33\%\\ 92.79\%\\ 91.30\% \\ 93.77\% \end{tabular}&
\begin{tabular}{@{}c@{}} \underline{90.00\%} \\ 89.00\%\\ \underline{90.00\%}\\ 90.00\% \\ \underline{90.00\%} \end{tabular}&
\begin{tabular}{@{}c@{}} \textbf{5.23\%} \\ \textbf{6.97\%}\\ \textbf{7.58\%}\\ \textbf{8.60\%} \\ \textbf{6.45\%} \end{tabular}\\

%---------------------------------------mnist---------------------------------------
\multicolumn{6}{c}{}\\
%------------------------------------f-mnist----------------------------------------
% \multicolumn{1}{c}{Defense}&
% \multicolumn{1}{c}{\begin{tabular}{@{}c@{}}Target \\ Classifier\end{tabular}} &
% \multicolumn{1}{c}{\begin{tabular}{@{}c@{}}No \\ Attack\end{tabular}} &
% \multicolumn{1}{c}{\begin{tabular}{@{}c@{}}FGSM\end{tabular}}&
% \multicolumn{1}{c}{\begin{tabular}{@{}c@{}}  PGD\\ $\epsilon$ = 0.3 \\ $\alpha$ = 0.01 \end{tabular}}& \multicolumn{1}{c}{\begin{tabular}{@{}c@{}}  C\&W\\ $\ell_{2}$ norm\end{tabular}} & \multicolumn{1}{c}{\begin{tabular}{@{}c@{}}\textbf{SPT}\end{tabular}} \\

\hline
\begin{tabular}{@{}c@{}}No \\ Defense\end{tabular}&
\begin{tabular}{@{}c@{}} $C_{p}$\\ $C_{a0}$ \\ $C_{a1}$\\ $C_{a2}$\\$C_{a3}$ \end{tabular}&
\begin{tabular}{@{}c@{}}$91.18\%^{*}$ \\ 91.35\% \\ 90.67\%\\ 89.79\%\\91.24\% \end{tabular}&
\begin{tabular}{@{}c@{}}$7.73\%^{*}$ \\ 11.19\% \\ 11.23\%\\ 15.67\%\\ 11.72\% \end{tabular}&
\begin{tabular}{@{}c@{}}$0.00\%^{*}$ \\ \textbf{0.04\%} \\ \textbf{0.58\%}\\ \textbf{3.01\%}\\ \textbf{1.68\%} \end{tabular}&
\begin{tabular}{@{}c@{}}$0.00\%^{*}$ \\ 88.00\% \\ 90.00\%\\ 89.00\%\\ 90.00\% \end{tabular}&
\begin{tabular}{@{}c@{}}$ 10.00\% ^{*}$\\ \underline{10.02\%} \\ \underline{10.00\%}\\ \underline{13.36\%}\\ \underline{10.00\%}  \end{tabular}\\

% \begin{tabular}{@{}c@{}} \\  99.91\%\\ 100.00\%\\81.65\% \\ 100.00\% \end{tabular}\\

% \hline
% \begin{tabular}{@{}c@{}}Defense- \\GAN-Rec \\ $L = 200$ \\ $R = 10 $\end{tabular}&
% \begin{tabular}{@{}c@{}} p\\ a0 \\ a1\\ a2\\a3 \end{tabular}&
% \begin{tabular}{@{}c@{}}91.18\% \\ 91.35\% \\ 90.67\%\\ 89.79\%\\91.24\% \end{tabular}&&&&\\

\hline 
\begin{tabular}{@{}c@{}}  PGD Adv. Tr.\\(Tr. Acc) \\ $\epsilon$ = 0.3  \\ $\alpha$ = 0.01 \end{tabular}&
\begin{tabular}{@{}c@{}} $C_{p}$\\ $C_{a0}$ \\$C_{a1}$\\ $C_{a2}$\\$C_{a3}$ \end{tabular}&
\begin{tabular}{@{}c@{}}74.80\% \\ 72.40\% \\ 73.27\%\\ 77.29\%\\71.47\% \end{tabular}&
\begin{tabular}{@{}c@{}}\underline{71.65\%} \\ \underline{67.58\%} \\ \underline{68.54\%}\\ \underline{72.77\%}\\69.50\% \end{tabular}&
\begin{tabular}{@{}c@{}} 71.69\% \\ 67.78\%\\ 68.69\%\\73.28\% \\\underline{69.36\%} \end{tabular}&
\begin{tabular}{@{}c@{}} 90.00\% \\ 90.00\%\\ 90.00\%\\90.00\% \\90.00\% \end{tabular}&
\begin{tabular}{@{}c@{}} \textbf{9.99\%} \\ \textbf{10.00\%}\\ \textbf{9.99\%}\\ \textbf{10.00\%} \\\textbf{10.00\%} \end{tabular}\\
% \begin{tabular}{@{}c@{}}  99.98\%\\ 100.00\%\\ 99.99\%\\  99.93\%\\  100.00\%\end{tabular}\\
\hline

\end{tabular}
\vspace{-4mm}
\end{table*}
% 1. The trained adversaries can transfer across different target model without loss successful attack rate.\\

In the section, we present experimental results %of SPT comparing with other attacks under 
in the black-box setting. The same five target networks as in the white-box setting are used. Among them, Classifier $C_{p}$ %^ {*}$ %\red{What's the difference between Classifier$_{p}^*$ and Classifier$_{p}$?} \blue{same one. * denote substitute model.} 
is used as the substitute model. %and train it on all the training images. 
We first generate adversarial examples against $C_{p}$ %^{*}$ %similar to white-box attacks, 
and then test them on the other four target networks. As expected, SPT consistently obtains low accuracy with or without defense. Thus, it has excellent transferability and is extremely effective in the black-box setting. FGSM and PGD have rather low accuracy when no defense is applied due to their moderate transferability. However, when defense is applied, both of them exhibit poor transferability. %and cannot effectively attack other target models in the black-box setting. 
C\&W exhibits poor transferability in all the scenarios due to the excessive optimization applied~\cite{kurakin2016adversarial}. %However, the C\&W exhibits bad transferability due to the excessive optimization\cite{kurakin2016adversarial}. 
% In addition, as the "class-persevering proportion" shown in  Table \ref{tab:white_attack} and \ref{tab:black_attack}, we find that the SPT adversarial examples even classified by other target models(rather than the substitute used to generate these adversarial examples), the predicted class often same as that by substitute. The property is what targeted attack expected, even we don't focus on it in this paper. 
%Combined the result shown in white-box experiment together, our SPT steadily obtain low accuracy under both white-box and black-box attacks whether defense is applied or not. 
Further,
% Table~\ref{tab:mnist_challenge} shows that 
SPT reduces the accuracy of the black-box Madry network to $9.80\%$ in the MNIST challenge, while the current best result on the leaderboard is $92.76\%$~\cite{xiao2018generating}.

\subsection{Illustration of Adversarial Examples}
%In the section, we present the generated adversaries on MNIST and fashion-MNIST dataset. We demonstrate our adversarial examples are natural and diverse. 
% Both MNIST and fashion-MNIST provide an intuitive definition of what is natural and what is diverse. i.e., Do the generated digit images look like something a person would write? Are the generated clothes images like the clothes we usually see? Whether the generated digit images appear to draw from same generator or not? Are the generated clothes images the same one or not? 
% We train five target CNN classifiers to generate adversaries against, and their architectures can be found in appendix. We treat all these classifiers as white-boxes, and present the generated adversaries in \ref{fig:example} with examples of each original image.
Figure~\ref{fig:example} shows the adversarial examples generated by SPT where the first row gives the original images and each other row shows the generated adversarial examples against $C_p$, $C_{a0}, C_{a1}, C_{a2}$, and $C_{a3}$, respectively. We observe that adversarial examples on the same column (corresponding to different target models) 
have distinct colors (brightness) but they all keep the same structure as the original image. In fact, SPT generates diverse adversarial examples not only for different target classifiers but also for different initialization settings (see Figure 4
%~\ref{fig:init_example} 
 in the 
%Appendix
technical report~\cite{SPT}
).

%\red{different initialization is not shown in the figure; also, is this true for other attack approaches?} \blue{added in appendix}
The diversity of adversarial examples poses a challenge to defense against SPT as it is difficult to resist all kinds of SPT adversarial examples. %The defense cannot resist SPT attack, only if they can resist all of kind of SPT adversarial examples.
Moreover, all the adversarial examples are clean %, realistic and meaningful 
and legible to humans. For MNIST, the generated digits keep the same handwriting characteristics, %while the color (brightness) of background and digit changed, which are
like the same person writing with different-colored pens and papers. 
For F-MNIST, the generated clothes are of the same style while they have different colors and textures. %which seem to be a variety of clothes of the same theme.

%\subsection{Analysis and Rethinking DNNs}
\subsection{Observations and Discussions}
We make the following observations from the above results.  
%find some interesting phenomenons of SPT adversarial examples against DNNs. 
%First DNNs can achieve high accuracy on classifying original images while exhibiting low accuracy on classifying SPT adversarial examples as shown in Tables\ref{tab:white_attack} and \ref{tab:black_attack}. % The reason why deep learning can achieve high precision on original test images is they r
%This is mainly because DNNs rely on the {\it i.i.d} assumption.DNNs select the hypothesis function from hypothesis space by learning training data, so that the hypothesis function correctly classify data which obeying identical distribution with training dataset. However, 

First, DNNs even with defense are vulnerable to adversarial examples %have limited ability to correctly the data not 
drawn from a distribution that is different from the training dataset. %And our adversarial examples do not obey same distribution with original images. 
That is the main reason why the proposed SPT adversarial examples exhibit high attack success rate and high transferability. %whether defense be applied or not. 

Second, as shown in Figure~\ref{fig:example}, %for a column, there are different adversarial examples for same one original images. 
adversarial examples in the same column preserve the same structure pattern, but they are classified into  different classes by the target DNNs. This indicates that DNNs are not very good at learning shapes and structures especially when images are not drawn from the same distribution as the training data. 

Third, Figure~\ref{fig:example} shows that adversarial examples on the same row are often classified into the same class. The statistics of the prediction results across labels given in Table 3
%~\ref{tab:pred_statistics}
% (in the Appendix)
 in our detailed report~\cite{SPT}
 further confirms this observation. We note that these examples correspond to different original images against the same target model. Moreover, they exhibit different structures but have similar brightness. %We found they often are classified as same one class, 
%For data drawing from a different distribution with training data, 
% DNNs do not learn the variety between pixels (i.e. structure) and 
% \blue{the predicted classes by DNNs mainly depend on the input feature space (varies with brightness) but not the underlying shapes contained in the images.}
We therefore speculate 
that in make predictions for image drawing from different distribution with training data, %the predicted classes by 
DNNs mainly rely on the distribution of pixel values rather than detailed structure patterns. 
%interpixel variety.

We hope our observations bring new insights into how DNNs work. %learn.

% These may interpret the phenomenon mentioned above and the high success attack rate we will show in the following attack evaluation. Further, The adversaries can transfer across different target model with retaining predicted category with high probability as shown in Table \ref{}. That provide more powerful evidence for the speculation above. 

\section{Related Works}

% \blue{Some recent attacks also use more structured perturbations beyond simple norm bounds. For example, \cite{sharif2016accessorize} shows that wearing eyeglass frames can cause face-recognition models to misclassify.
% \cite{fawzi2016measuring} tests the robustness of classifiers to "nuisamce variables", such as geometric distortions, occlusions, and illumination changes of the images. 
% There are some designed perturbation hid in graffiti \cite{evtimov2017robust} and stickers \cite{brown2017adversarial}. 
% % both \cite{zhao2017generating} and \cite{song2018generative} generate adversarial examples through GAN. 
% In \cite{zhao2017generating}, the researchers proposes mapping the input image to a latent space using GANs, and search for adversarial examples in the vicinity of the latent code. Similarity,  \cite{song2018generative} propose mapping noise vector with target class to adversarial example using GANs, by searching for optimal noise vector. And in  \cite{song2018generative}, the researchers also redefine a general adversarial example, while the adversarial example need label ground-truth manually.}

There are a few existing works that consider generating general adversarial example beyond the small perturbation requirement. Most of them are built upon the idea that the ground truth of an original image can propagate to adversarial examples by preserving the visual similarity between them. %original image and the adversarial examples. 
In~\cite{zhao2017generating,song2018generative}, %the visual similarity %between adversarial and original images 
this is achieved by limiting the change of a latent representation of images learned using GANs. %between two images through GAN. 
Another approach %preserve visual similarity by 
is to make perturbations resemble real yet inconspicuous objects to hide them from humans~\cite{evtimov2017robust,brown2017adversarial}. 
% \red{I don't understand this.}
% moreover, in \cite{sharif2016accessorize}, researchers impersonate the adversarial perturbation into eyeglass frame in face recognition task. 
%----0905---------
%In addition, researchers \cite{song2018generative} generate adversarial example to match ground truth by using AC-GAN \cite{odena2016conditional} which can incorporate label information and generate labeled image.

The intuition behind the recent work~\cite{hosseini2018semantic} is somewhat similar to ours, where they propose the concept of ``shape bias" and state that humans prefer to categorize objects according to their shapes rather than colors~\cite{landau1988importance}. However, the technical approaches of the two works are very different. In particular, the approach in ~\cite{hosseini2018semantic} focuses on color images and the main idea is convert the color space from RGB to HSV. To preserve the shape information, they keep the brightness component unchanged and change color components (i.e., saturation and hue) with the same amount for all the pixels. In contrast, we change pixel values based on structure patterns, where the amount of change is different for pixels in different structure patterns. Moreover, our transformation is controlled by multiple trainable parameters. These two differences are the key to obtain more diverse  adversarial examples compared to the approach in ~\cite{hosseini2018semantic}. We also note that although our approach focuses on grey-scale images, the key idea applies to color images as well.

\section{Conclusion and Further Work}
In this paper, we propose the technique of structure-preserving transformation (SPT) to generate natural and diverse adversarial examples with high transferability. %from original images beyond small perturbation requirement. 
SPT keeps the semantic similarity between adversarial examples and original images by preserving the structure patterns between them. This new approach allows the ground truth of original images to be shared with adversarial examples without imposing the small perturbation requirement.   
Empirical results on the MNIST and fashion-MNIST datasets show that SPT can generate adversarial examples that are natural to humans while being sufficiently diverse.  
Further, the adversarial examples significantly reduce the classification accuracy of target models %to very low accuracy
whether defenses are applied or not. In particular, we show that SPT can easily bypass PGD-based adversarial training. Moreover, SPT adversarial examples can transfer across different models with little or no loss of successful attack rates. The high successful attack rates and outstanding transferability stem from the key property that SPT adversarial examples follow different distributions from the training data of target models. 
% Finally, we show that SPT can easily bypass the PGD adversarial training in the public MNIST challenge, reducing the accuracy of the white-box network to 9.79\% and the black-box network to 9.80\%.
% from the best result in learder board with 88.79\% and dropping the accuracy of black-box networks to 9.80\% while the best result is 92.76\%.

Although we focus on untargeted attacks and gray-scale images in the paper, the core idea of SPT can extend to targeted attacks and color images.  In our preliminary experiments for color images, the high successful attack rate is at the expense of images being less  discriminable and authentic.
How to ensure the quality of images %under the premise of 
while maintaining a high successful attack rate for color images is an important challenge that we will investigate in our future work. In addition, we will test the performance of SPT-based adversarial training and study the performance of SPT attacks under other defense methods that do not rely on $\ell_p$ attacks.  

%In this paper, we focus on untargeted attacks and gray-scale images. However, the core idea of SPT readily extends to targeted attacks and color images, which will be investigated in our future work.

%\bigskip
%\noindent Thank you for reading these instructions carefully. We look forward to receiving your electronic files!
% \clearpage
% \newpage
%\red{check references. make sure the format is consistent. replace CoRR by arXiv link} 
%\blue{corrected, please check if they make sense.}
%\newpage{~~}
%\newpage{}
\bibliography{ref}
\bibliographystyle{aaai}
%\newpage{~~}
%\newpage{}
% \input{appendix.tex}
\end{document}